\pdfoutput=1
\documentclass[11pt]{article}

\usepackage{acl}

\usepackage{times}
\usepackage{latexsym}

\usepackage[T1]{fontenc}
\usepackage[utf8]{inputenc}

\usepackage{microtype}

\usepackage{enumitem}
\usepackage{booktabs}
\usepackage{graphicx}
\usepackage{multirow}
\usepackage{float}
\usepackage{adjustbox}

\usepackage{hyperref}
\usepackage[english]{babel}
\addto\extrasenglish{%

}

\definecolor{Green}{RGB}{76, 119, 59}

\newcommand{\guillemet}[1]{``#1''}

\title{QFrCoLA: a Quebec-French Corpus of Linguistic Acceptability Judgments}

\author{David Beauchemin \and Richard Khoury\\
    Group for Research in Artificial Intelligence of Laval University (GRAIL)\\
    Université Laval, Québec, Canada \\\{\texttt{david.beauchemin, richard.khoury}\}\texttt{@ift.ulaval.ca}
    }

\begin{document}
\maketitle
\begin{abstract}
Large language models (LLM) perform outstandingly in various downstream tasks. 
However, there is limited understanding regarding how these models internalize linguistic knowledge, so various linguistic benchmarks have recently been proposed to facilitate syntactic evaluation of language models (LM) across languages.
This paper introduces QFrCoLA (Quebec-French Corpus of Linguistic Acceptability Judgments), a normative binary acceptability judgments dataset comprising 25,153 in-domain and 2,675 out-of-domain sentences.
Our study leverages the QFrCoLA dataset and seven other linguistic binary acceptability judgments corpus to benchmark eight LM.
The results demonstrate that, on average, fine-tuned Transformer-based LM are strong baselines for most languages and that zero-shot binary classification LLM perform worse than the naive baseline on the task.
However, for the QFrCoLA benchmark, on average, a fine-tuned Transformer-based LM outperformed other methods tested.
It also shows that pre-trained cross-lingual LLMs selected for our experimentation do not seem to have acquired linguistic judgment capabilities during their pre-training for Quebec French.
Finally, our experiment results on QFrCoLA show that our dataset, built from examples that illustrate linguistic norms rather than speakers' feelings, is similar to linguistic acceptability judgment; it is a challenging dataset that can benchmark LM on their linguistic judgment capabilities.
\end{abstract}

\section{Introduction}
\label{sec:intro}

The introduction of large language models (LLM) \cite{touvron2023llama} and Transformer-based language model (LM) \cite{vaswani2017attention} has led to significant progress in natural language processing (NLP), substantially increasing the performance of most NLP tasks \cite{zhang2023instruction}. 
LLMs were initially introduced for English \cite{kenton2019bert, brown2020language, touvron2023llama}, but many other languages were later introduced, such as Russian \cite{kuratov2019adaptation}, French \cite{martin2020camembert}, and Norwegian \cite{kummervold-etal-2021-operationalizing}. 
NLP research has approached the competencies evaluation of various natural language tasks of LM with various benchmark corpora such as the English benchmarks GLUE \cite{wang-etal-2018-glue}, SuperGLUE \cite{wang2019superglue}, and GLGE \cite{liu2021glge} to name a few.
These corpora are collections of resources for training, evaluating, and analyzing LM \cite{eval-harness, chang2023survey}.
For example, GLUE aims to benchmark an NLP system's capabilities for natural language understanding (NLU) \cite{wang-etal-2018-glue}. At the same time, GLGE focuses on natural language generation (NLG) tasks such as document summarization \cite{liu2021glge}. 

Recently, much effort has been put into creating linguistic acceptability resources to assess and benchmark LM linguistic competency, where recent NLP research formulate linguistic competency as a binary classification task \cite{cherniavskii-etal-2022-acceptability, proskurina2023can}. 
That is the ability, from a native speaker's perspective, to distinguish the correct form and naturalness of an acceptable sentence from an unacceptable one \cite{chomsky2014aspects}.
Recently, similar non-English resources have been proposed to answer this question in typologically diverse languages such as Japanese \cite{someya2023jcola}, Norwegian \cite{jentoft2023nocola}, and Chinese \cite{hu2023revisiting}. However, the ability of LMs to perform linguistic acceptability judgments in French remains understudied.

To this end, we introduce the \textbf{Q}uebec-\textbf{Fr}ench \textbf{C}orpus \textbf{o}f \textbf{L}inguistic \textbf{A}cceptability Judgments (QFrCoLA), a corpus consisting of 25,153 in-domain and 2,675 out-of-domain normative acceptability judgment sentences, making it the second largest linguistic acceptability resources available in the NLP literature.
The main contributions of this work are therefore
\begin{enumerate}[leftmargin=*, noitemsep, topsep=0ex]
    \item The creation and release of \href{https://github.com/GRAAL-Research/QFrCoLA}{QFrCoLA}\footnote{\href{https://github.com/GRAAL-Research/QFrCoLA}{https://github.com/GRAAL-Research/QFrCoLA}}, a dataset of normative grammatical and ungrammatical sentences with binary labels; 
    \item A set of experiments to assess the performance of LM on QFrCoLA; 
    \item A cross-lingual benchmarking of LM on eight languages, including French, that opens up novel multi-language research perspectives.
\end{enumerate}

It is outlined as follows: first, we study the available linguistic binary acceptability corpus and related binary classification LM research in \autoref{sec:rel_work}. 
Then, we propose the QFrCoLA in \autoref{sec:frcola}, and in \autoref{sec:experiment} and \autoref{sec:res} we present a set of experiments
aimed at testing the performance of LM binary classifiers on all the linguistic acceptability resource corpora. 
Finally, in \autoref{sec:conclusion}, we conclude and discuss our future work.

\section{Related Work}
\label{sec:rel_work} 
Linguistic acceptability judgment evaluates one capacity to distinguish the correct form and naturalness of an acceptable sentence from an unacceptable one. 
For instance, individuals can inherently distinguish between two sentences and identify the one that is more acceptable or natural-sounding. This assessment is the primary behavioural benchmark employed by generative linguists to investigate the underlying structure of human language \cite{chomsky2014aspects}. 
Through benchmarking linguistic acceptability judgments of LLM, one can assess their linguistic robustness.

\subsection{Language Model Evaluation}
Historically, evaluation of LMs has been conducted using metrics or benchmark corpora \cite{chang2023survey}. 
The first approach relies either on task-agnostic metrics, such as perplexity \cite{jelinek1977perplexity} which measures the quality of the probability distribution of words in a given corpus by a model, or on task-specific metrics, like the BLEU score that evaluates a model’s performance for machine translation \cite{Papineni02bleu:a}.
The second approach relies on large corpora designed for NLU or NLG downstream tasks. For example, the GLUE benchmark \cite{wang-etal-2018-glue} is used to assess a model's NLU performance on tasks such as semantic similarity, linguistic acceptability judgment and sentiment analysis. 
In contrast, GLGE \cite{liu2021glge} evaluates language generation tasks such as summarization and question answering.

\subsection{Language Model Linguistic Acceptability Judgments Evaluation}
Recently, NLP researchers started using linguistic acceptability judgment tasks to assess the robustness of LMs against grammatical errors \cite{9968310} and to probe their grammatical knowledge \cite{choshen-etal-2022-grammar, mikhailov2022rucola}.
Two approaches are used to perform this evaluation: minimal pairs and binary classification acceptability judgments \cite{chang2023survey}.

In the first approach, a set of minimal pairs of grammatically acceptable and unacceptable sentences, such as the pair illustrated in \autoref{tab:blimp}, is presented to an LM. By observing which sentences the LM assigns a higher correctness probability to, one can assess which grammatical phenomena it is sensitive to \cite{warstadt2019linguistic}. 
Corpus such as BLiMP in English \cite{warstadt2019linguistic} and CLiMP in Chinese \cite{xiang2021climp} have been proposed to enable the evaluation of LM on a wide range of linguistic phenomena.

\begin{table}
    \centering
    \footnotesize
    \begin{tabular}{l|l}
    Acceptable Sentence & Not Acceptable Sentence         \\\midrule
    The cats annoy Tim.     & The cats annoys Tim.     
    \end{tabular}%
    \caption{Example of a minimal pair \cite{warstadt2019linguistic}.}
    \label{tab:blimp}
    \vspace{-1em}
\end{table}

In the second approach, a set of sentences that are either grammatical or ungrammatical, such as the two shown in \autoref{tab:itacola}, are provided to an LM which must perform a binary classification \cite{warstadt2019neural}.
Seven corpora have been proposed to assess LMs' capabilities to discriminate proper grammar from improper in their respective languages: CoLA for English \cite{warstadt2019neural}, DaLAJ for Swedish \cite{volodina2021dalaj}, ITACoLA for Italian \cite{trotta2021monolingual}, RuCoLA for Russian \cite{mikhailov2022rucola}, CoLAC for Chinese \cite{hu2023revisiting}, NoCoLA for Norwegian \cite{jentoft2023nocola} and JCoLa for Japanese \cite{someya2023jcola}.
However, as of yet, no such corpus exists for French.

Typically, the datasets in the second approach comprise sentences collected from syntax textbooks and linguistics journals. 
These datasets propose \guillemet{in-domain} train-dev-test splits to train and evaluate machine learning models.
CoLA, RuCoLA, CoLAC, and JCoLA corpora also include an \guillemet{out-of-domain} (OOD) split to assess whether a model suffers from overfitting. 
However, the definition of OOD varies depending on the corpus. 
CoLA includes sources of varying degrees of domain specificity and time period compared to those used for the primary dataset \cite{warstadt2019linguistic}. 
For RuCoLA, they are sentences generated by an automatic machine translation system and paraphrase generation models and annotated by a human annotator \cite{mikhailov2022rucola}.
While JCoLA comprises sentences from the Journal of East Asian Linguistics, a source with typically more complex linguistic phenomena than the other reference use of the in-domain splits \cite{someya2023jcola}.

\begin{table}
    \centering
    \resizebox{0.49\textwidth}{!}{%
    \begin{tabular}{cl}
    \toprule
    Label & \multicolumn{1}{c}{Sentence}           \\\midrule
    \texttt{0} (Ungrammatical)    & Edoardo returned to his last year city \\
    \texttt{1} (Grammatical)   & This woman has impressed me          \\\bottomrule 
    \end{tabular}%
    }
    \caption{Example sentences from the ItaCoLA dataset \cite{trotta2021monolingual}.}
    \label{tab:itacola}
    \vspace{-1em}
\end{table}

\section{QFrCoLA: Quebec-French Corpus of Linguistic Acceptability Judgments}
\label{sec:frcola}
In this work, we introduce the \textbf{Q}uebec-\textbf{Fr}ench \textbf{C}orpus \textbf{o}f \textbf{L}inguistic \textbf{A}cceptability Judgments (QFrCoLA), which will be the first large-scale normative binary linguistic acceptability judgments dataset for the Quebec-French language and the second-largest corpus in any language.

\subsection{Sources}
QFrCoLA consists of French normative grammatical or ungrammatical sentences taken from two online French sources: the \guillemet{\textit{Banque de dépannage linguistique}} (BDL) and the \textit{Académie française}. The first source is our \guillemet{in-domain} Quebec-French sentences for the train-dev-test splits, while the second is our OOD hold-out split.
Both sources are publicly available online, and we obtained authorization to publish them under a CC-BY-NC 4.0 license.

\subsubsection{In-Domain Source}
\begin{figure}
    \centering
    \includegraphics[width=\linewidth]{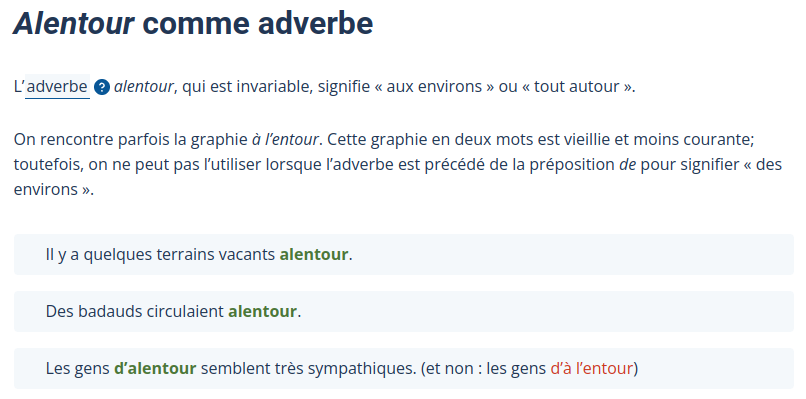}
    \caption{Snipped of the BDL article for the French adverb \guillemet{\textit{alentour}}. The text is in French.}
    \label{fig:bdl}
    \vspace{-1em}
\end{figure}

The BDL is an official online resource created by the \guillemet{\textit{Office québécois de la langue française}} (OQLF), a provincial government public organization in Canada\footnote{The Quebec government created the OQLF to \guillemet{protect} the French Quebec culture \cite{molinari2008anglais, bobowska2009quebec}, therefore it can be considered as a \guillemet{political initiative}. \cite{dahlet2010orthographe}. Consequently, its BDL initiative can be perceived as a biased French grammatical resource. However, the accepted grammar of the BDL is similar to other French native communities such as Belgium and Switzerland \cite{saint2013attitudes}.}, making it a reliable normative French resource.
It is a normative grammatical resource of 2,667 articles divided into eleven categories, such as \guillemet{\textit{orthographe}} (spelling), and \guillemet{\textit{syntaxe}} (syntax).
These articles explain various normative linguistic phenomena that the OQLF considers correct or incorrect. 
It uses examples written by French linguists to illustrate both cases based on linguistic phenomenal observation. 
For example, the \guillemet{\textit{adverbes}} (adverbs) category includes an article about the linguistic phenomenon of proper and improper use of the adverb \guillemet{\textit{alentour})} (surrounding). 
\autoref{fig:bdl} displays examples of well-written sentences using the adverb (in \textcolor{Green}{green}) and an example of an erroneous usage (in \textcolor{red}{red}). 

\subsubsection{Out-Of-Domain Source}
\begin{figure}
    \centering
    \includegraphics[width=\linewidth]{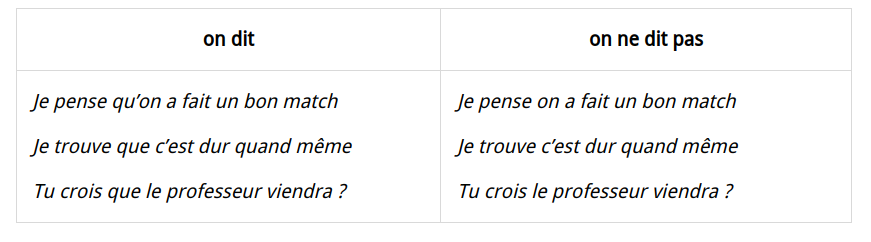}
    \caption{Snipped of an \textit{Académie française} article for the \guillemet{\textit{Omission de la conjonction « que »} (Omission of the conjunction "that")}. The text is in French.}
    \label{fig:academie}
    \vspace{-1em}
\end{figure}

Our second source is the \textit{Académie française}, a France-based organization acting as a \guillemet{society of scholars} in science and literature \cite{academie}.
It publishes monthly in their online \textit{La langue française: Dire, Ne pas dire} journal that presents 1,013 articles on normative grammar with examples of proper and improper use of French.
These examples are sorted into three categories: \guillemet{\textit{néologismes and anglicismes}} (neologisms and anglicisms), \guillemet{\textit{emplois fautifs}} (wrongful employment), and \guillemet{\textit{extensions de sens abusives}} (abusive extensions of meaning).
\autoref{fig:academie} displays examples of proper (left) and wrongful (right) 
employments of the conjunction \guillemet{\textit{que}} (that).

Like CoLA, RuCoLA, CoLAC, and JCoLA, our corpus includes an OOD split using a similar approach as JCoLA and CoLA.
Namely, we use a substantially different source to build it. Indeed, French in Quebec differs from France \cite{fagyal2006french}. For example, the feminization of titles differs between the two; the feminization of \textit{auteur} (author) in Quebec is accepted as \textit{autrice} or \textit{auteure} \cite{auteur}, while in France it is only accepted as \textit{auteure} \cite{auteuraf}.
However, both countries have similar linguistic phenomena, such as syntax and plurals \cite{dankova2017storytelling}.

\subsection{Data Collection}
\subsubsection{In-Domain}
We examined all 2,667 articles and manually extracted 25,153 normative linguistic acceptability judgment sentences.
Each sentence was labelled \texttt{0} (ungrammatical) or \texttt{1} (grammatical) following the BDL \textcolor{Green}{green}/\textcolor{red}{red} colour scheme as illustrated in \autoref{fig:bdl}.
Furthermore, since the BDL uses a fine-grained category structure to sort various linguistic phenomena, we collected these categories and associated them to labels according to the French linguistic literature \cite{fagyal2006french, chesley2010lexical, boivin2020categorisation, feldhausen2021revisiting}, and labelled each extracted sentence accordingly.
Our linguistic phenomena labels are listed below, and \autoref{tab:categorization} presents QFrCoLA statistics for each one, along with an example. 
Our categories are unevenly distributed, with nearly 43\% being in the morphology category. 
Moreover, the percentage of acceptability labels is also unevenly distributed, ranging from 58.26\% to 77.56\%.
It is due to the nature of our dataset, where the BDL, in many cases, presents proper normative use of French rather than improper use. It is shown for the \guillemet{anglicism} where nearly every sentence presents a proper and improper case.

\begin{table*}
\resizebox{\textwidth}{!}{%
        \begin{tabular}{lp{0.6\textwidth}cl}
        \toprule
        Category& BDL Fine-Grained Categories &\begin{tabular}[c]{@{}l@{}}\# Sen \\ \% Acp\end{tabular} &  Example\\
        \midrule
        Syntax   & Agreement violations, corruption of word order, misconstruction of syntactic clauses and phrases, incorrect use of appositions, violations of verb transitivity or argument structure, ellipsis, missing grammatical constituencies or words     & \begin{tabular}[c]{@{}c@{}} 5,152 \\ 77.24
        \end{tabular}    & \begin{tabular}[c]{@{}l@{}}\textit{Dès son arrivée, on s’empressa de lui poser des questions à propos de son voyage.}\\ (translated) As soon as he arrived, people were quick to ask him questions about his trip.\\\textbf{\textit{\underline{Dès en arrivant}, on s’empressa de lui poser des questions à propos de son voyage.}}\\
        (translated) \textbf{\underline{As soon as he arrived}, they were quick to ask him questions about his trip.}
        \end{tabular}         \\ \midrule
        Morphology & Incorrect derivation or word building, non-existent words    & \begin{tabular}[c]{@{}c@{}} 10,642 \\ 68.26 
        \end{tabular}  & \begin{tabular}[c]{@{}l@{}}
        \textit{Sa maison est neuve.}\\
        (translated) His house is new.\\
        \textbf{\textit{Sa maison est \underline{neuf}.}}\\
        (translated) \textbf{His house is \underline{new}.}
        \end{tabular} \\\midrule
        Semantic  & Incorrect use of negation or violates the verb’s semantic argument structure    & \begin{tabular}[c]{@{}c@{}} 5,442 \\ 72.97
        \end{tabular}        &                                                                         \begin{tabular}[c]{@{}l@{}}
        \textit{Quand la parade est passée, le vieil homme s’est levé pour aller voir à la fenêtre.}\\
        (translated) When the parade was over, the old man got up to look out the window.\\
        \textbf{\textit{Quand la parade est passée, le vieil homme \underline{s’est levé debout} pour aller voir à la fenêtre.}}\\
        (translated) \textbf{When the parade passed, the old man \underline{stood up} to look out the window.}
        \end{tabular}                                                  \\\midrule
        Anglicism  & Word and syntactical structure borrowed from English grammar   & \begin{tabular}[c]{@{}c@{}} 3,917 \\ 57.18
        \end{tabular}       &  \begin{tabular}[c]{@{}l@{}}
        \textit{Sauront-Ils répondre aux les besoins de l’enfant?}\\
        (translated) Will they be able to meet the child's needs?\\
       \textbf{\textit{Sauront-Ils \underline{rencontrer} les besoins de l’enfant?}}\\
       (translated)  \textbf{Will they be able to \underline{meet} the child's needs?}
        \end{tabular}
        \\\bottomrule                                                      
    \end{tabular}%
    }
    \caption{Number of sentences (\# Sen) and the percentage of acceptable sentences (\% Acp) per category in QFrCoLA (all three splits), and example of a positive and a negative (\textbf{bolded} with error \underline{underlined}) along with their translation in each category.}
    \label{tab:categorization}
\end{table*}

\subsubsection{Out-Of-Domain}
OOD sentences were manually extracted from the journal's 1,013 articles.
We extracted 2,675 sentences from those articles and only binary labelled them following the table scheme (left/right) as illustrated in \autoref{fig:academie}. We discuss the dataset statistics in the following section.

\subsection{Comparison With Other Similar Corpora}
This section compares our corpus with all related ones. \autoref{tab:stats1} present in-domain number of sentences, percentage of acceptable sentences and vocabulary size for the train, dev and test sets\footnote{It is worth mentioning that for CoLA, RuCoLA and JCoLA, their in-domain test set labels are not available to reduce the risk of overfitting. Thus, like other related work \cite{cherniavskii-etal-2022-acceptability}, we use their out-of-domain dev sets as the test sets. Also, CoLAC does not provide an OOD set nor label for their test set. Thus, per the authors' recommendation, the in-domain train and dev set was resampled using a 60-10-30\% split with seed 42 to create new splits.} and for the entire corpus.
The total vocabulary sizes were computed using language-specific SpaCy tokenizers \cite{Honnibal_spaCy_Industrial-strength_Natural_2020} that split each sentence into individual words or punctuation.
We can see that QFrCoLA is the second largest corpus in terms of the number of sentences it contains, behind only NoCoLA, and is approximately twice the size of all the other corpora.
Moreover, it has a similar frequency of acceptable sentences to the CoLA, CoLAC, and RuCoLA datasets, and like the other corpora, all splits have a similar frequency of acceptable sentences.
Finally, we can see that QFrCoLA has the third-largest vocabulary size compared to the other datasets.

\begin{table*}
    \centering
    \resizebox{\textwidth}{!}{%
    \begin{tabular}{lcccccccccc|ccc}
    \toprule
    & \multirow{2}{*}{Language} & \multicolumn{3}{c}{Train} & \multicolumn{3}{c}{Dev} & \multicolumn{3}{c|}{OOD/Test}                                                    &     \multicolumn{3}{c}{Total}   \\
    & & \multicolumn{1}{c}{\# Sen} & \multicolumn{1}{c}{\% Acp} & Vocab & \multicolumn{1}{c}{\# Sen} & \multicolumn{1}{c}{\% Acp} & Vocab & \multicolumn{1}{c}{\# Sen} & \multicolumn{1}{c}{\% Acp} & \multicolumn{1}{c|}{Vocab} & \multicolumn{1}{c}{\# Sen} & \multicolumn{1}{c}{\% Acp} & Vocab \\\midrule
 
    CoLA \cite{warstadt2019neural} & English  & 8,551 & 70.44 & 5,778 & 527  & 69.26 & 1,375 & 516                 & 68.60 & 988 & 9,594 & 70.27 & 6,097 \\
    DaLAJ \cite{volodina2021dalaj} & Swedish & 7,682   & 50.00 & 6,841 & 890 & 50.00 & 1,799   & 888 & 50.00     &1,661             & 9,460 &50.00 & 7,884 \\
    ITACoLA \cite{trotta2021monolingual} & Italian & 7,801  & 84.39 & 5,825 & 946 & 85.41 & 1,844  & 1,888 & 84.21  & 1,888             & 9,722 & 84.47 & 6,402\\
    RuCoLA  \cite{mikhailov2022rucola} & Russian & 7,869 & 74.52 & 19,057 & 983  &74.57 & 4,140  & 1,804 & 63.69 & 9,353                & 10,656 & 72.69 & 26,382\\
    CoLAC \cite{hu2023revisiting}  & Chinese & 4,134 & 66.09 & 3,835 & 460 &66.96 & 1,024  & 1,970         &67.82   & 2,636  & 6,564 & 66.67 & 4,759\\
    NoCoLA \cite{jentoft2023nocola} & Norwegian & 116,195 &31.46 & 32,561 & 14,289 & 32.59 & 8,865 & 14,383 & 31.58  & 8,600           & 144,867  & 31.58 & 37,319\\
    JCoLA \cite{someya2023jcola} & Japanese & 6,919 & 83.38 & 3,730 & 865  & 83.93 & 1,483 & 684 & 73.28 & 896               & 8,469   & 82.62 & 4,146\\\midrule
    QFrCoLA  & French & 15,846 & 69.49 & 18,350 & 1,761 & 69.51 & 5,369  & 7,546 & 69.49  &12,690           & 25,153  & 69.49&22,131\\\bottomrule
    \end{tabular}%
    }
    \caption{Comparison of QFrCoLA and related corpora for the number of sentences (\# Sen), percentage of acceptable sentences (\% Acp), and vocabulary size (Vocab). \guillemet{OOD} stands for \guillemet{out-of-domain}.}
    \label{tab:stats1}
\end{table*}

\autoref{tab:stats3} present, for the OOD split, the number of sentences, vocabulary size and percentage of acceptable sentences of all linguistic corpora with an available OOD split.
However, since other corpora do not distribute their hold-out labels, we could not compute the percentage of acceptable sentences.
We also note that for JCoLA, the OOD hold-out split was unavailable in their \href{https://github.com/osekilab/JCoLA}{official dataset GitHub repository}.
Once again, we can see that QFrCoLA is the second largest corpus in terms of number of sentences and vocabulary size, with nearly as many sentences as RuCoLA.
Compared to the main QFrCoLA corpus in \autoref{tab:stats1}, we can see that the OOD split comprises a much less diverse vocabulary, making it well distinct from the other splits.
Finally, the OOD hold-out split has a percentage of acceptable sentences nearly 15\% lower than the overall corpus, making it more robust to highlight overfitting cases in machine learning models.

\begin{table}[ht!]
    \footnotesize
    \centering
    \begin{tabular}{lccc}
    \toprule
    & \multicolumn{3}{c}{OOD Hold-Out} \\
    & \multicolumn{1}{c}{\# Sen} & \multicolumn{1}{c}{Vocab} & \multicolumn{1}{c}{\% Acp}\\\midrule
    CoLA & 533                & 1035 & N/A\\
    RuCoLA  & 2,789              & 12,211 & N/A\\
    CoLAC  & 931          & 1,168& N/A\\
    JCoLA & N/A              &  N/A & N/A\\\midrule
    QFrCoLA   & 2,675           &  1,651& 53.91\\\bottomrule
    \end{tabular}%
    \caption{Comparison of QFrCoLA with all related corpus with an out-of-domain (OOD) hold-out set for the number of sentences (\# Sen), the vocabulary size (Vocab) and the \% of acceptable sentences (\% Acp).}
    \label{tab:stats3}
\end{table}

\section{Experiments}
\label{sec:experiment}
We train and evaluate three fine-tuned approaches and evaluate eight LLMs in a zero-shot binary classification setup. We then benchmark these models against a baseline.

\subsection{Evaluation Metrics}
Following \citet{warstadt2019neural}, performance is measured using the accuracy score and Matthews correlation coefficient (MCC) \cite{matthews1975comparison}.
Accuracy on the dev set is used as the target metric for hyperparameter tuning and early stopping. 
We report the results averaged over ten restarts from different random seeds (i.e. $[42, 43, \cdots, 50, 51]$).

\subsection{Models}
As our baseline, we selected the trivial approach to always select class \texttt{1} (\texttt{Baseline}). 
Namely, this model accuracy equals the percentage of acceptable sentences (\% Acp) illustrated in \autoref{tab:categorization}.

\subsubsection{Monolingual Language Model}
\label{ref:baseline}
\paragraph{Monolingual}
We selected a state-of-the-art (SOTA) pre-trained monolingual LM for each language based on their benchmark performance on various tasks \cite{chang2023survey} as our monolingual baseline (\texttt{BERT}).
We detail the selected language-specific model name in \autoref{tab:models}.

\begin{table}
    \resizebox{0.5\textwidth}{!}{%
    \begin{tabular}{lp{0.5\textwidth}}
    \toprule
    Language       & Model Name                                \\\midrule
    En          & bert-base-cased \cite{kenton2019bert}  \\
    SV         & bert-base-swedish-cased \cite{swedish-bert} \\
    IT       & bert-base-Instructtalian-cased \cite{stefan_schweter_2020_4263142} \\
    RU        & ruBert-base \cite{zmitrovich2023family}\\
    ZH         & bert-base-chinese \cite{cui2021pre} \\
    NO        & nb-bert-base \cite{kummervold-etal-2021-operationalizing} \\
    JA         & bert-base-japanese \cite{bert-japan} \\
    FR        & camembert-base \cite{martin2020camembert} \\\bottomrule                 
    \end{tabular}%
    }
    \caption{Selected pre-trained transformer models per language using ISO-2 letter format.}
    \label{tab:models}
\end{table}

\paragraph{State-Of-The-Art}
The SOTA approach to binary linguistic acceptability judgments is the topological data analysis (TDA) proposed by \citet{cherniavskii-etal-2022-acceptability} (\texttt{LA-TDA}).
This approach extracts the attention maps of a fine-tuned Transformers-based LM to use as linguistic features to train a binary logistic regression. 
The authors report that this approach significantly outperformed previous approaches, increasing the MCC score on linguistic acceptability for English, Italian, and Swedish by up to $0.24$. 
In our case, we use the attention maps from the monolingual fine-tuned models.
We selected this approach since it is the SOTA approach.

\subsection{Cross-Lingual Language Model}
To assess whether cross-lingual LM approaches can benefit from using linguistic phenomena from various languages, we compare a Transformer-based cross-lingual baseline against four cross-lingual LLMs.
Our objective is to evaluate cross-lingual LM linguistic capabilities across various languages.

\paragraph{Fine-Tuned Transformer-Based Cross-Lingual Language Model}
For our cross-lingual baseline, we use XLM-\texttt{RoBERTa}-base \cite{conneau-etal-2020-unsupervised}, a Transformer-based approach.

\paragraph{Zero-Shot Large Language Model}
Benchmarking all available LLM was outside the scope of this article due to a lack of resources to process the evaluation.
LLM benchmark articles have reported using many SOTA GPU devices to do such evaluation \cite{kew2023bless}, which we do not have at our disposal.
We instead selected five LLMs that were 1) open-source, 2) around 7B parameters, and 3) have been shown to perform well on various benchmark \cite{kew2023bless, xu2023superclue, malode2024benchmarking}, or optimized for generation of French text, namely \texttt{BLOOM}-7B \cite{le2023bloom}, \texttt{BLOOMZ}-7B \cite{yong2022bloom}, \texttt{Mistral}-7B-v0.3 \cite{jiang2023mistral}, \texttt{LLama}-3.1-8B \cite{dubey2024llama}, and \texttt{Lucie} \cite{openllm2023claire} (optimized for French) along with their instruct variants (\texttt{I}), if available.
We benchmarked all LLMs using HuggingFace's \texttt{zero-shot-classification}.

\subsection{Training Settings}
Each \texttt{BERT} LM is fine-tuned using the language-specific train and dev split, while \texttt{RoBERTa} LM uses all the languages train and dev splits. 
All models are evaluated using the test or, if available, OOD split following the standard procedure under the HuggingFace library \cite{wolf-etal-2020-transformers}.
Each model is fine-tuned for four epochs and uses the AdamW optimizer \cite{loshchilov2018decoupled}, with a learning rate of $3e{-5}$ and a weights decay of $1e{-2}$.
Since the corpora are unbalanced, we use a weighted balanced loss based on the train split percentage of acceptable sentences.
We use a batch size of 32 and the HuggingFace default train hyperparameters.
For each LM, we use the default tokenizer with a maximum sequence length of 64 tokens without lowercasing during tokenization.

\section{Results and Discussion}
\label{sec:res}
\subsection{In-domain Results}
\autoref{tab:res1} presents the accuracy and the MCC of all models for each benchmark dataset on the dev and test sets, with \textbf{bolded} value indicating the best score per benchmark.
Except for the zero-shot evaluation setup, the table reports the average and one standard deviation over the ten restarts.
We observe that, for most languages, on average \texttt{LA-TDA} outperforms other fine-tuned methods, but not on all metrics and with a smaller margin than reported by \citet{cherniavskii-etal-2022-acceptability}. 
The two exceptions to this are CoLA and QFrCoLA. 
QFrCoLA performs slightly better using the fine-tuned \texttt{BERT} model.
Considering that \texttt{LA-TDA} is computed asymptotically in quadratic time \cite{cherniavskii-etal-2022-acceptability}, the performance gains seem marginal compared to the added computational expense. 
These results show that fine-tuned Transformer-based LM are strong baselines for the binary linguistic acceptability classification tasks.

Moreover, LLM accuracy performances are either worse than the baselines or at par with it for all languages except Norwegian. 
In the case of Norwegian, performance is slightly better than the baseline.
\texttt{Llama} achieves the worst performance across all languages; however, \texttt{BLOOMZ} and \texttt{Mistral} perform best for most languages.
We also observed that, for all LLMs, the instruct (\texttt{I}) version of the LLM performs better than the non-instruct one by, for most of them, a large margin (i.e. double or less the performance).
Furthermore, all LLM achieve poor MCC on all splits, with scores close to 0, meaning a negligible correlation between the prediction and the labels. 
Our experimentation results show that pre-trained cross-lingual LLMs selected for our experimentation do not seem to have acquired linguistic judgment capabilities during their pre-training, nor French optimized LLM (\texttt{Lucie}). Indeed, we can see that even Lucie performed poorly on the task, with an accuracy below the naive approach. Moreover, even our fine-tuned approach (\texttt{RoBERTa}) does not seem to acquire cross-lingual linguistic capabilities from potentially similar linguistic phenomena amongst languages. It shows that leveraging multilingual linguistic corpus to train a multilingual acceptability judgment LM is complex, and more work needs to be done to achieve better performance than the monolingual approach.
Most tested languages do not share a common grammatical language or alphabet (e.g., Japanese and Italian). Thus, it highlights that training LMs on a multilingual dataset without proper grammar assessment could lead to LMs not fully comprehending language linguistics.

Finally, our experiment results on QFrCoLA show that our dataset, which is built from examples that illustrate linguistic norms rather than speakers' feelings, is similar to linguistic acceptability judgment; namely, it is a challenging dataset that can be used to benchmark LM on their linguistic judgment capabilities.

\begin{table}
    \begin{adjustbox}{totalheight=\textheight-3\baselineskip, center, width=0.49\textwidth}
        \centering
        \begin{tabular}{lcccc}
        \toprule
        \multirow{2}{*}{Model} & \multicolumn{2}{c}{Dev}                            & \multicolumn{2}{c}{Test/OOD}                           \\
                                                                                                & Acc (\%) ($\uparrow$)                & MCC ($\uparrow$)                     & Acc (\%) ($\uparrow$)                 & MCC  ($\uparrow$)                    \\\midrule
        \multicolumn{5}{c}{CoLA}                                                                                                                                                                          \\\midrule
        \texttt{Baseline} & 69.26 & 0.000 & 68.60 & 0.000 \\
        \texttt{BERT}                                                                               & $ 83.61 \pm 2.56$          & $\mathbf{ 0.639 \pm 0.030}$ & $\mathbf{ 80.89 \pm 1.15}$ & $\mathbf{ 0.544 \pm 0.025}$ \\
        \texttt{LA-TDA}                                                                                     & $\mathbf{ 84.91 \pm 1.24}$ & $ 0.633 \pm 0.031$          & $ 80.70 \pm 1.38$          & $ 0.532 \pm 0.034$          \\
        \texttt{RoBERTa}                                                                              & $ 82.24 \pm 1.35$          & $ 0.575 \pm 0.033$          & $ 77.25 \pm 2.42$          & $ 0.452 \pm 0.041$    \\
        \midrule
        \texttt{BLOOM} & 31.88 & 0.019 & 32.56 & 0.051\\
        \texttt{BLOOMZ} & 64.14 & 0.151 & 60.47 & 0.044 \\
        \texttt{Mistral} & 30.93  & -0.039  & 33.72 & 0.073 \\
        \texttt{Mistral-I} & 63.57 & 0.005 & 62.02 & -0.043
        \\
        \texttt{Llama} & 55.03 & -0.003 & 58.53 & 0.021 
        \\
        \texttt{Llama-I} & 56.93 & -0.003 & 52.71 & -0.039
        \\

        \midrule\toprule
        \multicolumn{5}{c}{DaLAJ}                                                                                                                                                                         \\\midrule
        \texttt{Baseline} & 50.00 & 0.000 & 50.00 & 0.000 \\
        \texttt{BERT}                                                                               & $ 69.12 \pm 1.53$          & $\mathbf{ 0.411 \pm 0.029}$ & $ 72.33 \pm 1.40$          & $ 0.467 \pm 0.025$          \\
        \texttt{LA-TDA}                                                                                     & $\mathbf{ 70.08 \pm 1.24}$ & $\mathbf{ 0.411 \pm 0.024}$ & $\mathbf{ 73.54 \pm 1.05}$ & $\mathbf{ 0.475 \pm 0.020}$ \\
        \texttt{RoBERTa}                                                                              & $ 55.18 \pm 5.90$          & $ 0.131 \pm 0.144$          & $ 55.21 \pm 5.89$          & $ 0.124 \pm 0.137$          \\
        \midrule
        \texttt{BLOOM} & 50.45 & 0.010 & 49.21  & -0.020 \\
        \texttt{BLOOMZ} & 50.90 & 0.047 & 49.77 & -0.011 \\
        \texttt{Mistral} & 65.52  & -0.016 & 66.63 & -0.014\\
        \texttt{Mistral-I} & 52.17  & -0.072 & 51.05 & -0.093
        \\
        \texttt{Llama} & 38.46 & -0.068 & 37.22 & -0.075
        \\
        \texttt{Llama-I} & 61.89 & 0.009 & 62.57 & 0.009
        \\

        \midrule\toprule
        \multicolumn{5}{c}{ITACoLA}                                                                                                                                                                       \\\midrule
        \texttt{Baseline} & 85.41 & 0.000 & 84.21 & 0.000 \\
        \texttt{BERT}                                                                               & $ 83.29 \pm 3.71$          & $ 0.420 \pm 0.051$          & $ 83.45 \pm 3.34$          & $\mathbf{ 0.446 \pm 0.050}$ \\
        \texttt{LA-TDA}                                                                                     & $\mathbf{ 87.51 \pm 0.88}$ & $\mathbf{ 0.423 \pm 0.050}$ & $\mathbf{ 86.59 \pm 0.93}$ & $ 0.422 \pm 0.054$  \\
        \texttt{RoBERTa}                                                                              & $ 79.97 \pm 6.22$          & $ 0.105 \pm 0.121$          & $ 79.12 \pm 5.99$          & $ 0.117 \pm 0.124$          \\\midrule
        \texttt{BLOOM} & 73.15 & 0.006 & 69.00 & -0.095 \\
        \texttt{BLOOMZ} & 54.97  & -0.058  & 55.28  & -0.052  \\
        \texttt{Mistral} & 15.33  & 0.036 &  16.72 & -0.014\\
        \texttt{Mistral-I} & 63.53 & -0.036 & 58.87 & -0.032
        \\
        \texttt{Llama} & 37.32 &  0.010 & 34.26 & -0.044
        \\
        \texttt{Llama-I} & 32.77 & -0.012 & 30.46 & -0.071
        \\

        \midrule\toprule
        \multicolumn{5}{c}{RuCoLA}   
        \\\midrule
        \texttt{Baseline} & 74.57 & 0.000 & 63.69 & 0.000 \\
        \texttt{BERT}                                                                               & $ 74.49 \pm 2.56$          & $\mathbf{ 0.352 \pm 0.027}$ & $ 66.81 \pm 3.56$          & $ 0.379 \pm 0.030$          \\
        \texttt{LA-TDA}                                                                                     & $\mathbf{ 77.56 \pm 0.61}$ & $ 0.337 \pm 0.022$          & $\mathbf{ 71.09 \pm 0.92}$ & $\mathbf{ 0.382 \pm 0.018}$ \\
        \texttt{RoBERTa}                                                                              & $ 71.84 \pm 3.00$          & $ 0.276 \pm 0.038$          & $ 56.81 \pm 3.18$          & $ 0.189 \pm 0.026$          \\\midrule
        \texttt{BLOOM} & 37.44 & -0.084 & 47.56 & -0.012 \\
        \texttt{BLOOMZ} & 59.91 & 0.014 & 51.05 & -0.040 \\
        \texttt{Mistral} & 26.25 & 0.036 & 36.97 & 0.014 \\
        \texttt{Mistral-I} & 61.65  & -0.052 & 58.76 & -0.055
        \\
        \texttt{Llama} & 61.95 & 0.028 & 53.10 & 0.049
        \\
        \texttt{Llama-I} & 34.99 & 0.008 & 44.57 & -0.037
        \\

        \midrule\toprule
        \multicolumn{5}{c}{CoLAC}                                                                                    \\\midrule
        \texttt{Baseline} & 66.96 & 0.000 & 67.82 & 0.000 \\
        \texttt{BERT}                                                                               & $ 75.93 \pm 1.35$          & $ 0.444 \pm 0.027$          & $ 77.78 \pm 1.43$          & $ 0.482 \pm 0.023$          \\
        \texttt{LA-TDA}                                                                                     & $\mathbf{ 77.33 \pm 1.79}$ & $\mathbf{ 0.469 \pm 0.044}$ & $\mathbf{ 79.01 \pm 0.86}$ & $\mathbf{ 0.502 \pm 0.023}$ \\
        \texttt{RoBERTa}                                                                              & $ 73.37 \pm 2.72$          & $ 0.337 \pm 0.022$          & $ 71.09 \pm 0.92$          & $ 0.382 \pm 0.018$          \\\midrule
        \texttt{BLOOM} & 66.96 & 0.000 & 67.71 & 0.001 \\
        \texttt{BLOOMZ} & 63.91 & -0.029 & 65.03 & -0.015 \\
        \texttt{Mistral} & 32.83 & -0.064 & 33.15 & 0.005 \\
        \texttt{Mistral-I} & 38.91 & -0.003 & 37.41 & -0.016\\
        \texttt{Llama} & 62.61 &  -0.040 & 64.67 & -0.007\\
        \texttt{Llama-I} & 63.48 & 0.026 & 63.76 & 0.005\\

        \midrule\toprule
        
        \multicolumn{5}{c}{NoCoLA}                                                                                   \\\midrule
        \texttt{Baseline} & 32.59 & 0.000 & 31.58 & 0.000 \\
    
        \texttt{BERT}                                                                               & $ 77.90 \pm 0.96$          & $ 0.560 \pm 0.009$          & $ 77.90 \pm 0.98$          & $ 0.560 \pm 0.009$          \\
        \texttt{LA-TDA}                                                                                     & $\mathbf{ 81.58 \pm 0.29}$ & $\mathbf{ 0.582 \pm 0.007}$ & $\mathbf{ 82.01 \pm 0.31}$ & $\mathbf{ 0.589 \pm 0.009}$ \\
        \texttt{RoBERTa}                                                                              & $ 73.92 \pm 1.40$          & $ 0.504 \pm 0.017$          & $ 73.79 \pm 1.37$          & $ 0.505 \pm 0.015$          \\\midrule
        \texttt{BLOOM} & 61.10 & 0.013 & 61.31 & 0.003 \\
        \texttt{BLOOMZ} & 35.92 & -0.047 & 36.92 & -0.033 \\
        \texttt{Mistral} & 65.52 & -0.016 & 66.63 & -0.014\\
        \texttt{Mistral-I} & 52.17 & -0.072 & 51.05 & -0.093\\
        \texttt{Llama} & 38.46 & -0.068 & 37.22 & -0.075\\
        \texttt{Llama-I} & 61.89 & 0.009 & 62.57 & 0.009\\

        \midrule\toprule
        \multicolumn{5}{c}{JCoLA}                                                                                   \\\midrule
        \texttt{Baseline} & 83.93 & 0.000 & 73.28 & 0.000 \\
        \texttt{BERT}                                                                               & $ 81.34 \pm 4.48$          & $ 0.039 \pm 0.062$          & $ 73.17 \pm 0.61$          & $ 0.067 \pm 0.111$          \\
        \texttt{LA-TDA}                                                                                     & $\mathbf{83.49 \pm 0.68}$ & $ 0.252 \pm 0.051$          & $\mathbf{ 75.30 \pm 1.25}$ & $ 0.230 \pm 0.070$          \\
        \texttt{RoBERTa}                                                                              & $ 72.64 \pm 8.11$          & $\mathbf{ 0.262 \pm 0.058}$ & $ 72.86 \pm 4.61$          & $\mathbf{ 0.328 \pm 0.059}$ \\\midrule
        \texttt{BLOOM} & 24.51 & 0.036 & 31.82 & 0.000 \\
        \texttt{BLOOMZ} & 81.39  & -0.002 & 70.22 & -0.007 \\
        \texttt{Mistral} & 18.84 & 0.031 & 29.05 & 0.054 \\
        \texttt{Mistral-I} & 25.09 & -0.016 & 33.43 & 0.035\\
        \texttt{Llama} & 31.33 & 0.006 & 36.64 & 0.000  \\
        \texttt{Llama-I} & 62.54 & 0.001 & 56.20 & -0.126\\

        \midrule\toprule
        \multicolumn{5}{c}{QFrCoLA}                                                                                                                                                                        \\\midrule
        \texttt{Baseline} & 69.51 & 0.000 & 69.49 & 0.000 \\
        \texttt{BERT}                                                                               & $\mathbf{ 84.51 \pm 0.78}$ & $\mathbf{ 0.619 \pm 0.02}$  & $\mathbf{ 82.92 \pm 0.61}$ & $\mathbf{ 0.578 \pm 0.015}$ \\
        \texttt{LA-TDA}                                                                                     & $ 84.00 \pm 0.48$          & $ 0.606 \pm 0.013$          & $ 82.79 \pm 0.45$          & $ 0.574 \pm 0.012$       \\
        \texttt{RoBERTa}                                                                              & $70.67 \pm 15.13$         & $ 0.243 \pm 0.263$          & $69.91 \pm 14.61$         & $ 0.222 \pm 0.240$          \\\midrule
        \texttt{BLOOM} & 32.71 & 0.007 & 32.94 & 0.020 \\
        \texttt{BLOOMZ} & 64.00 & 0.043 & 61.75 & -0.011 \\
        \texttt{Mistral} & 33.50 & -0.005 & 33.45 & -0.002 \\
        \texttt{Mistral-I} & 63.03 & -0.020 & 63.61 & -0.007\\
        \texttt{Llama} & 45.43 & -0.019 & 45.44 & 0.000  \\
        \texttt{Llama-I} & 46.45 & -0.026 & 48.25 & -0.001\\
        \texttt{Lucie} & 60.14 & 0.041 & 58.18 & -0.008\\
        \texttt{Lucie-I} & 36.40 & -0.034 & 38.87 & 0.011\\

        \bottomrule   
        \end{tabular}%
    \end{adjustbox}
    \caption{Acceptability binary classification results and MCC by language. The best score per benchmark is \textbf{bolded}. \guillemet{OOD} stands for \guillemet{out-of-domain}. $\uparrow$ means higher is better}
    \label{tab:res1}
\end{table}

\subsection{Out-Of-Domain Results}
We present in \autoref{tab:cat} the accuracy and the MCC of our three models trained using QFrCoLA over the dataset's four categories along with the six LLM evaluated in a zero-shot binary classification setup. 
Except for the LLM, the table reports the average and one standard deviation over the ten restarts.
We can see that the category \guillemet{anglicism} has the lowest performance for the Transformer-based LM.
For the two approaches using monolingual LLM (i.e. \texttt{BERT} and \texttt{LA-TDA}), we hypothesize that this situation is due to occurrences of anglicism in the LLM training dataset.
Indeed, using word and syntactical structure borrowed from English grammar is more common over web-based \cite{laviosa2010corpus, planchon2019anglicisms, solano2021anglicisms, vsukaliccorpus} and even official educational text \cite{simon2021use}.
Thus, fine-tuning the pre-trained LLM model can be more challenging, considering that the \guillemet{anglicism} category contains the least examples.
For the cross-lingual approach, since the LLM has learned word representation over English during training, we hypothesize that sentences using English words or syntax are considered more probable for the model; thus, it is more challenging for the classifier to classify these examples correctly.
For the LLM, the \guillemet{anglicism} performances are worse than the other category and the baseline.

Our experimentation results show that pre-trained cross-lingual or French optimized LMs selected for our experimentation do not seem to have acquired linguistic judgment capabilities during their pre-training, even on the more dominant France-French. Indeed, France has more publicly available datasets online to train LM on, such as OSCAR \cite{2022arXiv220106642A}. 
It shows that these tested LMs do not seem to have acquired linguistic capabilities from their monolingual training nor from other languages.

Moreover, LLM accuracy performance is always worse for all categories than the baseline, and predictions correlate weakly with labels. It shows again that the benchmarked LLMs do not seem to have a linguistic understanding of Quebec French.

Finally, we present in \autoref{tab:res2} the accuracy and the MCC of our three models trained using QFrCoLA but evaluated using our OOD hold-out set. 
The table reports the average and one standard deviation over the ten restarts.
We can see that, once again, the \texttt{BERT} model outperforms the \texttt{LA-TDA} model. 
However, all three models show significant performance drops, of nearly 22\% in accuracy and nearly 50\% for the MCC. It shows that the fine-tuned models have overfitted over the train and dev dataset. 
As stated before, it is also worth noting that the French in Quebec differ from the French in France. 
These differences could explain the lower performance observed on the OOD split.

\begin{table}
    \resizebox{0.49\textwidth}{!}{%
    \begin{tabular}{lcccc}
    \toprule
    \multirow{2}{*}{Model} & \multicolumn{4}{c}{Category}                                                                              \\
                           & Syntax                   & Morphology               & Semantic                 & Anglicism                \\\midrule
    \multicolumn{5}{c}{Test Accuracy (\%) ($\uparrow$)}                                                                                             \\\midrule
    \texttt{Baseline} & 77.24 & 68.26  & 72.97  & 57.18 \\
    \texttt{BERT}         & $\mathbf{88.59 \pm 0.60}$  & $\mathbf{81.76 \pm 0.74}$  & $\mathbf{85.82 \pm 0.40}$  & $\mathbf{74.36 \pm 1.40}$  \\
    \texttt{LA-TDA}                 & $ 88.40 \pm 0.23 $           & $ 81.49 \pm 0.51 $           & $ 85.39 \pm 0.53 $           & $ 74.18 \pm 1.44 $           \\
    \texttt{RoBERTa}       & $ 83.31 \pm 4.31 $           & $ 74.93 \pm 4.70 $           & $ 79.84 \pm 4.88 $           & $ 63.79 \pm 4.66 $           \\\midrule
    \texttt{BLOOM} & 57.67 & 56.33 & 55.03 & 57.36\\
    \texttt{BLOOMZ} & 65.66  & 61.02 & 64.36 & 54.61 \\
    \texttt{Mistral} & 26.53  & 34.08 & 30.97 & 44.86 \\
    \texttt{Mistral-I} & 67.52 & 63.76 & 64.97 & 55.76
    \\
    \texttt{Llama} & 42.14 & 46.00 & 45.70 & 48.05
    \\
    \texttt{Llama-I} & 46.74 & 48.81 & 47.88 & 49.29
    \\
    \texttt{Lucie} & 59.78 & 59.27 & 58.06 &53.01
    \\
    \texttt{Lucie-I} & 33.38 & 40.92 & 35.15  & 40.92
    \\
    \midrule\toprule
    \multicolumn{5}{c}{Test MCC ($\uparrow$)}                                                                                                  \\\midrule
    \texttt{Baseline} & 0.000  &0.000  &0.000  &0.000  \\
    \texttt{BERT}         & $\mathbf{0.654 \pm 0.018}$ & $\mathbf{0.563 \pm 0.017}$ & $\mathbf{0.620 \pm 0.011}$ & $\mathbf{0.506 \pm 0.028}$ \\
    \texttt{LA-TDA}                 & $ 0.649 \pm 0.009 $          & $ 0.555 \pm 0.013 $          & $ 0.609 \pm 0.014 $          & $ 0.405 \pm 0.026 $  \\
    \texttt{RoBERTa}       & $ 0.403 \pm 0.279 $          & $ 0.327 \pm 0.226 $          & $ 0.378 \pm 0.261 $          & $ 0.223 \pm 0.156 $          \\\midrule
    \texttt{BLOOM} & -0.017 & 0.024 & 0.002 & 0.140\\
    \texttt{BLOOMZ} & -0.044 & -0.003 & -0.024 & 0.008 \\
    \texttt{Mistral} & 0.002 & -0.016 & -0.002 & 0.034\\
    \texttt{Mistral-I} & -0.084 & 0.006 & -0.011 & 0.029
    \\
    \texttt{Llama} & -0.062 & 0.016 & 0.017 & -0.004
    \\
    \texttt{Llama-I} & -0.032 & 0.017 & -0.007  & -0.005
    \\
    \texttt{Lucie} & -0.014 & 0.005 & -0.019 & -0.001
    \\
    \texttt{Lucie-I} & 0.009 & 0.010 & 0.027 & 0.015
    \\
    \bottomrule       
    \end{tabular}%
    }
    \caption{Acceptability binary classification results and MCC for QFrCoLA per category. The best score is \textbf{bolded}. $\uparrow$ means higher is better.}
    \label{tab:cat}
    \vspace{-1.5em}
\end{table}

\begin{table}[ht!]
    \centering
    \footnotesize
    \begin{tabular}{lccc}    
    \toprule
    & \multicolumn{2}{c}{OOD Hold-Out} \\
    & \multicolumn{1}{c}{Acc (\%) ($\uparrow$)} & \multicolumn{1}{c}{MCC ($\uparrow$)}\\\midrule
    \texttt{Baseline} & 53.91 & 0.000 \\
    \texttt{BERT} & $\mathbf{62.69 \pm 1.13}$ &  $\mathbf{0.286 \pm 0.020}$\\
    \texttt{LA-TDA} & $61.36 \pm 0.90$ & $0.090 \pm 0.019$\\
    \texttt{RoBERTa} & $55.99 \pm 4.36$ & $0.107 \pm 0.088$\\\midrule
    \texttt{BLOOM} & 45.42 & -0.048 \\
    \texttt{BLOOMZ} & 53.73 & 0.028 \\
    \texttt{Mistral} & 46.34 & -0.003 \\
    \texttt{Mistral-I} & 53.06 & 0.002
    \\
    \texttt{Llama} & 49.30 & 0.017
    \\
    \texttt{Llama-I} & 49.30 & -0.019
    \\
    \texttt{Lucie} & 51.61 & -0.018 
    \\
    \texttt{Lucie-I} & 47.55 & -0.022
    \\
    \bottomrule
    \end{tabular}%
    \caption{Acceptability binary classification result on the QFrCoLA out-of-domain (OOD) hold-out set. The best score per benchmark is \textbf{bolded}. $\uparrow$ means higher is better.}
    \label{tab:res2}
\end{table}

\section{Conclusion and Future Works}
\label{sec:conclusion}

This article introduced QFrCoLA, the Quebec-French Corpus of Linguistic Acceptability Judgments, a dataset comprising 25,153 in-domain and 2,675 OOD sentences annotated with binary acceptability manually extracted from two official online linguistic normative resources.
It is the first such corpus in French and the second-biggest one in any language. 
We have evaluated the linguistic performances of two monolingual and one cross-lingual fine-tuned Transformer-based LM approaches and four cross-lingual LLM on eight binary acceptability judgement datasets.

Our results demonstrated that Transformer-based LM achieves high results on the binary classification task and are strong baselines.
When fined-tuned on QFrCoLA, a Transformer-based LM even outperforms the SOTA \texttt{LA-TDA} method proposed by \citet{cherniavskii-etal-2022-acceptability}. It also shows that pre-trained cross-lingual LLMs selected for our experimentation do not seem to have acquired linguistic judgment capabilities during their pre-training for Quebec French. Finally, our experiment results on QFrCoLA show that our dataset, which is built from examples that illustrate linguistic norms rather than speakers' feelings, is similar to linguistic acceptability judgment; namely, it is a challenging dataset that can be used to benchmark LM on their linguistic judgment capabilities.

In our future works, we plan to extend the granularity of our dataset linguistic phenomena and generate the complementary grammatical or ungrammatical sentence of each sentence in the dataset to create the first French minimal pair benchmark dataset. 
Moreover, we would also like to explore the linguistic phenomena errors generated by the LLM qualitatively.

\section*{Limitations}
All the sentences in QFrCoLA have been extracted from official linguistic sources on theoretical syntax and normative grammar. 
Therefore, those sentences are guaranteed to be theoretically meaningful, making QFrCoLA a challenging dataset.
However, the categories extracted automatically from the official source are skewed.
Indeed, as shown in \autoref{tab:categorization}, nearly 42\% of the dataset comprises morphological linguistic phenomena.
This imbalance means overrepresenting morphology examples, which could provide an incomplete evaluation of a LM's ability to perform the task.
Moreover, as discussed, the dataset is based on the OQLF, a Quebec-French government organization, and the \textit{Académie française}; thus, the dataset represents normative grammar.
Furthermore, Quebec and France share a common grammar base but differ in some points, such as feminization (e.g. \textit{auteure}/\textit{autrice}). Thus, as discussed, the out-of-domain hold-out is a challenging split since it might represent accepted grammar use in Quebec rather than in France.

\section*{Ethical Considerations}
QFrCoLA may serve as training data for binary linguistic acceptability judgment classifiers \cite{batra-etal-2021-building}, which may benefit the quality of generated texts. We acknowledge that such text generation progress could lead to misusing LLMs for malicious purposes, such as disinformation or harmful text generation and online harassment \cite{weidinger2021ethical, bender2021dangers}. Nevertheless, our corpus can be used to train adversarial defence against such misuse and to train artificial text detection models \cite{lewis-white-2023-mitigating, kumar-etal-2023-mitigating}.

\section*{Acknowledgements}
This research was made possible thanks to the support of a Canadian insurance company, NSERC research grant RDCPJ 537198-18 and FRQNT doctoral research grant. We thank the reviewers for their comments regarding our work.
We also thank the \textit{Office québécois de la langue française} for their help regarding the curation of the corpus.

\bibliography{custom}
\bibliographystyle{acl_natbib}

\end{document}